\documentclass{article}
\usepackage{spconf,amsmath,graphicx}
\usepackage{amssymb}
\usepackage{multirow}
\usepackage{CJKutf8}
\usepackage{multirow}
\usepackage[table]{xcolor}

\usepackage{eqparbox}
\usepackage[skins]{tcolorbox}
\usepackage[normalem]{ulem}
\useunder{\uline}{\ul}{}

\definecolor{Alizarin}{rgb}{0.90, 0.1, 0.26}

\definecolor{Seeblue}{rgb}{0.28, 0.51, 0.95}

\title{Learning Comment Generation by leveraging User-Generated Data}
\name{Zhaojiang Lin, Genta Indra Winata, Pascale Fung\thanks{This work is partially funded by ITS/319/16FP of the Innovation Technology Commission, HKUST 16214415 \& 16248016 of Hong Kong Research Grants Council, and RDC 1718050-0 of EMOS.AI.}}
\address{
  Center for Artificial Intelligence Research (CAiRE)\\
  Department of Electronic and Computer Engineering\\
  The Hong Kong University of Science and Technology, Clear Water Bay, Hong Kong\\ 
  \tt \{zlinao,giwinata\}@ust.hk, pascale@ece.ust.hk
}
\begin{document}
%
\maketitle
\begin{abstract}
Existing models on open-domain comment generation are difficult to train, and they produce repetitive and uninteresting responses. The problem is due to multiple and contradictory responses from a single article, and by the rigidity of retrieval methods. To solve this problem, we propose a combined approach to retrieval and generation methods. We propose an attentive scorer to retrieve informative and relevant comments by leveraging user-generated data. Then, we use such comments, together with the article, as input for a sequence-to-sequence model with copy mechanism. We show the robustness of our model and how it can alleviate the aforementioned issue by using a large scale comment generation dataset. The result shows that the proposed generative model significantly outperforms strong baseline such as Seq2Seq with attention and Information Retrieval models by around 27 and 30 BLEU-1 points respectively.
\end{abstract}


\section{Introduction}
\label{sec:intro}
Commenting on online articles has been a popular method to collect the opinions and interests from the crowd. The quality of the comments often represents the level of users' engagement \cite{P18-2025}. However, not every comment is relevant, and often they contain inappropriate content such as abusive language \cite{park2017one}. Normally, upvote count is used by online forums to rank the quality of a comment according to users' preferences and bury the irrelevant comments.
The persuasiveness of the comments positively correlates by the number of votes given to them \cite{Wei2016IsTP}. Usually, popular comments receive many upvotes from users. The upvotes count in a comment indicates the level of attention from readers to the comment. The count is also useful to distinguish between relevant user opinions and undesired content such as spams including advertisement, double posting, or any offensive comments.

In recent years, automatic comment generation has become a prominent topic. The ability to comment on an article requires natural language understanding to conceptualize the idea of the article and provide a relevant response. Previous generative models suffer from the mode collapse issue where the models produce samples with extremely low variety \cite{Metz2016UnrolledGA}. In comment generation task, we also suffer the same issue when we train our models on articles together with all comments that have a huge variance.  As a result, the model is hard to converge, and generated comments are not meaningful. On the other hand, an Information Retrieval (IR) approach can pick comments from real users, but this approach is not scalable.

\begin{figure}[!t]
  \centering
  \includegraphics[width=1.0\linewidth]{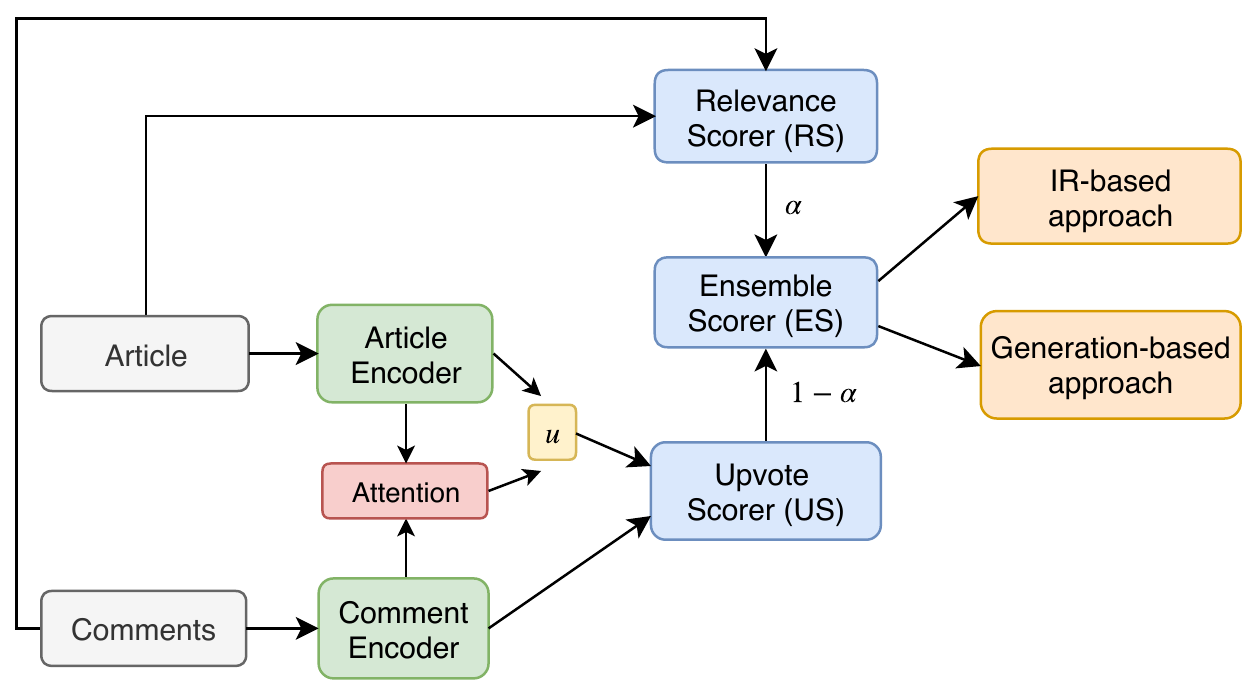}
  \caption{Comment Generation Framework}
  \label{fig:comment-generator}
\end{figure}

To build a scalable generator that can generate informative and relevant comments, we propose a framework to learn comment generation by leveraging user-generated data such as upvote count. Combining the user-generated data such as upvote count helps the model to decide which sequences are to be decoded. However, not every article has upvote information. To solve the problem, we propose to build a neural classifier to score the comments and combine with a generative model to effectively mitigate mode collapse issue. For the generation-based approach, we use a pointer-generator network \cite{see2017get} to learn and copy essential words from the articles. Our proposed method significantly improves the performance by around 25 BLEU-1 points, 3 CIDEr points, 5 ROUGE\_L points, and 6 METEOR points compared to standard pointer-generator network and it outperforms the IR models in most of the evaluation metrics.

\section{Related Work}
\label{sec:related-work}

Prior work in response generation focused on two major approaches: Information retrieval and Generative model.

\textbf{Information retrieval:} \cite{rao2018learning} proposed a neural-based method to clarify questions by calculating the expected value of the information in the technical discussion forum. In question answering, TF-IDF scores were employed to remove irrelevant answer candidates to reduce the search space \cite{P17-1171}.~A CNN encoder had been explored in automatic comment ranker to replace a standard IR-based similarity scorer \cite{P18-2025}. 

\textbf{Generative model:} ~\cite{8114175} introduced a gated attention neural-based generation model to address the problem of contextual relevance by choosing news context.~\cite{Wei2016IsTP} studied the impact of different sets of features to measure the comments' persuasiveness in the online forum. \cite{W17-3002} showed that argumentative comment representations are useful to identify constructive comments using dataset provided by \cite{napoles2017ynacc}. In summarization, the pointer-generator networks with copy mechanism were proposed by \cite{see2017get} to copy words from the source article to the generate a summary. Similarly to our work, the mechanism triggers the model to take some important keywords in the article and use it to generate relevant comments. \cite{P18-1136}\cite{wu2018end}\cite{wu2018globaltolocal} integrated copy mechanism with multi-hop memory network to efficiently utilize knowledge base in generating the response in task-oriented dialog system.

\section{Methodology}
\label{sec:methodology}

In this section, we describe a standard information retrieval (IR)-based method and generation-based method for comment generation. We propose a user-centered method for scoring comments by leveraging upvotes collected from users to improve the relevance of the generated comments. We denote $a \in \{w_1^a,...,w_n^a\}$ as the concatenation of the article and title with $n$ tokens, and $c \in \{w_1^c,...,w_m^c\}$ as the comment consists of $m$ tokens. As our proposed method, we apply pointer-generator network to encode and copy words from the article. Figure \ref{fig:comment-generator} depicts the pipeline of our comment generation architecture.

\subsection{Comment Scorer}
We propose three methods to score comments: a relevance scorer (RS), an upvote scorer (US), and an ensemble of both. 

\subsubsection{Relevance Scorer (RS)}
Similarly to IR-based methods, RS computes the dot product of TF-IDF weighted vector between an article with title $a$ and comment $c$. We normalize the scores with the maximum to the article.
\begin{equation}
S_r = \textnormal{Normalize}((v^{a})^{T}v^c)
\end{equation}
\subsubsection{Upvote Scorer (US)}
Here, we introduce a semi-supervised method to score comments in the articles without any upvotes. The model is trained on articles that have at least a comment with ten upvotes. Comments with ten upvotes or above are used as positive samples, and others as negative samples. First, we represent each word as an embedding vector and pass it to a bidirectional LSTM (BiLSTM). The model is shared for both articles with title and comments. 
\begin{equation}
\{h_1^a,...,h_n^a \} = \textnormal{BiLSTM}(\{v_1^a,...,v_n^a\}) 
\end{equation}
\begin{equation}
\{h_1^c,...,h_m^c \} = \textnormal{BiLSTM}(\{v_1^c,...,v_m^c\})
\end{equation} 

We compute attention scores \cite{luong2015effective} to capture the dependency between articles and comment.
\begin{equation}
u = \sum_{i=1}^{n}\frac{exp(e_{i})}{\sum_{k=1}^{n}exp(e_{k})}h_i^a
\end{equation}
where
\begin{equation}
e_{i} = (h_i^a)^T \textbf{W}_{a} h_{m}^c
\end{equation}
$e_{i}$ is the attention score and $\textbf{W}_a$ is a trainable parameter. We concatenate the context vector $u$ and comment feature vectors $h_m^{c}$ and pass it to a fully connected layer $\textnormal{F}$. Next, the resulting value is normalized with a sigmoid function.
\begin{equation}
\label{us}
S_u=\sigma(\textnormal{F}([u,h_m^c]))
\end{equation}
\begin{table*}[!ht]
\centering
\caption{\label{tab:table-name} Results on information retrieval and generation approaches. Higher scores are better.}
\begin{tabular}{l|l|l|l|l|l}
\hline
& \textbf{Model}                    & \textbf{BLEU-1} & \textbf{CIDEr} & \textbf{ROUGE\_L} & \textbf{METEOR} \\ \hline{}
\multirow{3}{*}{Retrieval} & TF-IDF + CNN \cite{P18-2025} & \textbf{35.55} & 0.25 & 21.92  & 14.25 \\    
& TF-IDF + RS & 34.67 & \textbf{1.37} & \textbf{23.67} & 14.80 \\  
& TF-IDF + ES & 34.53 & 1.19 & 23.59 & \textbf{14.85}  \\ \hline
\multirow{5}{*}{Generation} & Seq2Seq-Attn \cite{P18-2025} & 38.80 & 1.41 & 23.53 & 6.08 \\ 
& pointer-generator + coverage            & 40.84           & 1.29          & 25.44             & 6.49           \\
& pointer-generator + coverage + upvote     & 62.03           & 3.55          & 28.24    & 11.00           \\
& pointer-generator + coverage + RS & 56.39           & 3.95          & 26.47             & 11.88           \\  
& pointer-generator + coverage + US     & 64.22           & 4.17          & \textbf{30.87}    & 12.11           \\  
& pointer-generator + coverage + ES  & \textbf{65.70}  & \textbf{4.35} & 30.53 & \textbf{12.62} \\ \hline
\end{tabular}
\label{results}
\end{table*}

\subsubsection{Ensemble Scorer (ES)}

We combine the relevance score $S_r$ and upvote score $S_u$ by linear interpolation as an ensemble score. 
\begin{equation}
S = \alpha S_{r} + (1-\alpha)S_{u}
\end{equation}
where $\alpha \in [0,1]$ is a hyper-parameter to weight $S_{r}$ and $S_{u}$, We use the ensemble score to rank the comments by considering the relevance and user-generated upvote count.

\subsection{Information Retrieval-based Commenting}
This method is an unsupervised method to find the most relevant comment by retrieving comments from a large pool of candidates. In the standard IR-based method the articles and comments are represented as TF-IDF weighted bag-of-word vectors \cite{P18-2025}.  The tokens are bigrams, and we use the \textit{murmur3} hash function to map bigrams to $2^{24}$ space similar to \cite{P17-1171}. We apply a two-step retrieval to narrow down the search space and retrieve more relevant comment to the article. Given an article, we retrieve the top-$5$ most similar articles by calculating the dot product to their TF-IDF weighted vectors and build a candidate pool of comments from the retrieved articles.~Then, we apply three different methods to score the comments:

\begin{itemize}
\item As a baseline, we apply a convolutional neural network (CNN) \cite{lecun1998gradient} to encode articles and comments similar to \cite{P18-2025}. The inputs are tokens from an article and comments, and the model outputs the relevance score.
\item We use RS by taking the dot product of the TF-IDF weighted vector of articles and comments.
\item As our proposed method, we use ES and use them to rank the comments. 
\end{itemize}

Next, we rank them in descending order to their scores, and we choose a comment with the highest score.

\subsection{Generation-based Commenting}

We use Seq2Seq \cite{luong2015effective} with pointer-generator \cite{see2017get} that copy words from articles to generate comments. During decoding time, we calculate $p_{gen}$ $\in$ [0,1], the generation probability to weight vocabulary distribution for generation as following:
\begin{equation}
p_{gen} = \sigma (w_{h^*}^T h_t^* + w_s^T s_t + w_x^T x_t + b_{ptr})
\end{equation}
Next, the vocabulary distribution $P_{vocab}(w)$ is calculated by concatenating the decoder state $s_t$ and the context vector $h_t^*$. The final distribution $P(w)$ is calculated from the weighted sum of the attention weights $a$ and vocabulary distribution $P_{vocab}(w)$. To avoid repetitions, we added coverage \cite{P16-1008} in the loss function \cite{see2017get}. 
\begin{equation}
P(w) = p_{gen} P_{vocab}(w) + (1 - p_{gen})\sum_{i:w_i=w}{a_i^t}
\end{equation}
Similar to the IR-based approach, we use the proposed score functions to choose the best comments. We pair the best comment with the corresponding article to build the training set. Instead, for the standard seq2seq and pointer generator network with coverage baseline models, we create a set of training samples from all comments paired with the article. We also compare US with raw upvote as the score. We randomly sample one comment when there is an article without any upvote.

\begin{CJK*}{UTF8}{gbsn}
\begin{table*}[!t]
\caption{The generated comments from different models.}
\label{generated-comments}
\centering
\footnotesize
\begin{tabular}{|p{8cm}|p{2.2cm}|p{5.8cm}|}
\hline
\begin{tabular}[t]{@{}l@{}}\textbf{Title}:\enspace流浪狗被人残忍踢断下巴无法进食，路人心疼不已送它\\去医院。\\
(The stray dog was brutally kicked off the chin and could not eat. \\
A passerby sent it to the hospital.)
\end{tabular} & \textbf{Model} & \textbf{Generated Comment} \\ \hline
\multirow{5}{*}{\begin{tabular}[t]{@{}l@{}}
\textbf{Body}:\enspace一条流浪狗在路边的垃圾堆翻找着吃的东西，嘴里发出叫\\
声，似乎很痛苦，路过的男子见状将其抱回。他发现流浪\textbf{\textcolor{Seeblue}{狗狗}}\\
\textbf{\textcolor{Seeblue}{的下巴断了}}，似乎是人为踢断的，非常\textbf{\textcolor{Alizarin}{可怜}}，可以\textbf{\textcolor{Seeblue}{想象它有}}\\
\textbf{\textcolor{Seeblue}{多么的痛苦}}。狗狗很饿，在垃圾堆里找吃的，但是它的下巴变\\
成这样\textbf{\textcolor{Seeblue}{是无法吃东西的}}，不知道是什么人这么狠毒。这名男子\\
非常心疼，马上带\textbf{\textcolor{Alizarin}{狗狗}}去了宠物医院。经过医生的治疗，狗狗\\\
的下巴正在恢复中，现在能够吃东西了，不过只能吃一些流体\\
的、软的食物,回到咬骨头的时期恐怕是需要一段时间了。\\
(A stray dog was searching for food in the rubbish heap on the\\
roadside. The cry seemed to be very painful, and the passing man\\
saw it and took it home. He found \textbf{\textcolor{Seeblue}{the dog's chin was broken}}. It \\
seemed to be kicked by someone. Very \textbf{\textcolor{Alizarin}{poor}}. \textbf{\textcolor{Seeblue}{We can imagine how}} \\
\textbf{\textcolor{Seeblue}{painful it is}}. The dog was hungry, looking for food in the rubbish \\
heap, but \textbf{\textcolor{Seeblue}{it was unable to eat}} due to the broken chin. We don't know\\
who is so vicious. The man was very distressed, immediately took the\\
\textbf{\textcolor{Alizarin}{dog}} to the pet hospital. After the doctor's treatment, dog's jaw is\\
recovering, and now it can eat, but it can only eat some fluid, soft\\
food, it may take a while to go back to the time of biting bones.)
\end{tabular}} & TF-IDF~+~CNN~\cite{P18-2025} & \begin{tabular}[t]{@{}l@{}}
想起来我曾经养的一条小狗...后来, 因为\\
吃死老鼠被药死了。我一直都记着它那双\\ 
会说话的眼睛 。\\
(I remembered a puppy I used to raise... \\
Later, It was dead from food poisoning \\
because of a dead mouse. I always remember \\
its talking eyes.)
\end{tabular} \\
\cline{2-3} 
& TF-IDF~+~ES & 
\begin{tabular}[t]{@{}l@{}}听听都害怕, 踢断下巴\\
(I feel scared to hear that the chin was kicked off.)\\ 
\end{tabular} \\ 
\cline{2-3} 
& Seq2Seq-Attn &\begin{tabular}[t]{@{}l@{}}
好人有好报\\
(A good person will be rewarded) \\ \end{tabular} \\
\cline{2-3} 
& pointer-generator + coverage~+~RS& \begin{tabular}[t]{@{}l@{}}
\textbf{\textcolor{Seeblue}{狗狗的下巴断了}}，似乎\textbf{\textcolor{Seeblue}{是无法吃东西的}}，\\
但是只能\textbf{\textcolor{Seeblue}{想象它有多么的痛苦}}。\\
(\textbf{\textcolor{Seeblue}{The dog's chin was broken}}. It seemed that \textbf{\textcolor{Seeblue}{it}} \\
\textbf{\textcolor{Seeblue}{was unable to eat}}. But \textbf{\textcolor{Seeblue}{we can imagine how}} \\
\textbf{\textcolor{Seeblue}{painful it is}}.)\end{tabular} \\
\cline{2-3} & pointer-generator + coverage~+~ES &\begin{tabular}[t]{@{}l@{}} \textbf{\textcolor{Alizarin}{可怜}}的\textbf{\textcolor{Alizarin}{狗狗}}!\\(\textbf{\textcolor{Alizarin}{Poor dog}}!)\\ \end{tabular} \\ \hline
\end{tabular}
\end{table*}
\end{CJK*}

\section{Experiment}
\subsection{Dataset}
We collected a large-scale Chinese dataset from Tencent News, with four million real comments along with rich metadata using the script provided by \cite{P18-2025}. The dataset reported by \cite{P18-2025} has around 200K news articles and each of them has more than 20 comments. However, around 10\% of news articles are already expired. The dataset has 177,368 articles with four million users comments. Each article has a title, text body and metadata including user upvote count and article categories. Upvote is given by users, and it represents comment's popularity. Overall, the dataset has four upvotes per comments on average with a long-tail distribution, where most of the comments have zero upvotes. For popular comments, they have more than thousands of votes. The dataset is split into training/validation/test sets which contains $171,440/4,512/1,417$ samples respectively.

\subsection{Experimental Setup}

For all experiments, we use a bidirectional LSTM with a 256-dimensional hidden state for the encoder and a unidirectional LSTM with a 512-dimensional hidden state for the decoder. We use a word embedding with a size of 128 with a vocabulary size of 50k following \cite{see2017get}. The word embedding is shared with the encoder and decoder. We use Adam for the optimizer, and our model is trained with an initial learning rate of 1e-3, and it decays at the rate of 0.5 every epoch when there is an increase in validation perplexity. At the inference time, we replace all unknown tokens with the source tokens with the highest attention weight and block repeating tokens.

The upvote scorer is trained with the same setting as above, and we add a pre-trained 300-dimensional skip-gram Word2Vec \cite{mikolov2013distributed} trained on 374M words Sogou News\footnote{news.sogou.com} with 270,830 words vocabulary. The embedding layer is fixed in the entire training. We take 71,319 of 171,440 articles with at least one comment received ten upvotes or above as our training set. We compute the score for comments from the remaining 100,121 articles using equation (\ref{us}). The ensemble ratio $\alpha$ is set to 0.2. For the evaluation of models, we use the script from Coco Caption\footnote{https://github.com/tylin/coco-caption} \cite{chen2015microsoft}.

\begin{CJK*}{UTF8}{gbsn}
\section{Results}
\label{sec:results}

We show our empirical results in Table \ref{results}.~Overall, the generation-based approach outperforms IR in most of the metrics except METEOR. Adding a scorer in the IR-based model doesn't improve the retrieval performance. In contrast, the generation models such as pointer-generator networks receive advantages from using our proposed scorers. Our standard pointer-generator network achieves better performance regarding BLEU-1 and Rouge\_L by around 2 points. We achieve the state-of-the-art by combining ES to the pointer-generator model.

\textbf{Effectiveness of scorer:} Table \ref{generated-comments} is depicted to further analyze the results. We find that when we apply RS to the pointer-generator network, it starts to copy phrases from the articles and uses them to form a new comment. It shows that the model can preserve the relevance of the generated comments by utilizing RS.  Notably, using upvote scorer (US) is more effective than using raw upvote counts as the score, and it mitigates the issue when upvote information is not available in the unpopular articles.
Moreover, after applying ES, the generated comments are more expressive and solve the mode collapse issue in Seq2Seq with attention model that often generates common phrases such as ``我也是\enspace(me too)" and ``我也是醉了\enspace(I am speechless)".  

\textbf{Retrieval vs. Generation:} According to our observation, there are some trade-offs of choosing either IR or generation-based model. The comments from IR are always real human comments, and often they have very similar words to the articles. However, the candidates are very limited, and most of the time, they are irrelevant to the context. On the other hand, the sequences generated from the generation-based model learns how to decode important keywords and copies the words from the article text, and may produce the ungrammatical sequences. The proposed scorer helps the model to preserve the relevance and generate comments that are likely to be favored by humans. After applying the comment scorer, the performance increases significantly, and it can generate more relevant and attractive comments. 

\end{CJK*}

\section{Conclusion}
\label{sec:conclusion}
In this work, we present a novel framework for comment generation to leverage user-generated data and generate relevant comments according to the user preference.
Our results show that user-generated information helps the generative model. We efficiently alleviate the mode collapse issue by incorporating upvote scorer and relevance score to our model and produce more meaningful comments. For future work, we plan to apply reinforcement learning to the comment generation by taking scores as the reward.




\bibliographystyle{IEEEbib}
\bibliography{strings,refs}

\end{document}